\pdfoutput=1

\documentclass[11pt]{article}

\usepackage[]{EMNLP2023}

\usepackage{times}

\usepackage[T1]{fontenc}

\usepackage[utf8]{inputenc}

\usepackage{microtype}

\usepackage{inconsolata}
\usepackage{multirow}
\usepackage{graphicx}
\usepackage{booktabs}
\usepackage{amsmath}
\usepackage{amssymb}
\usepackage{bm}
\usepackage{caption}
\usepackage{subcaption}

\colorlet{plan_navigation}{DarkSeaGreen2}

\newcommand{\dataset}{TastyVidDial}
\newcommand{\modelname}[1]{MM-PlanLLM#1}

\title{Show and Guide: \\Instructional-Plan Grounded Vision and Language Model}

\author{Diogo Glória-Silva, David Semedo, João Magalhães \\
  NOVA LINCS, NOVA School of Science and Technology, Portugal \\
  \texttt{dmgc.silva@campus.fct.unl.pt}\\
  \texttt{\{df.semedo, jmag\}@fct.unl.pt
 }
}

\begin{document}
\maketitle
\begin{abstract}

Guiding users through complex procedural plans is an inherently multimodal task in which having visually illustrated plan steps is crucial to deliver an effective plan guidance.
However, existing works on plan-following language models (LMs) often are not capable of multimodal input and output. In this work, we present \modelname{}, the first multimodal LLM designed to assist users in executing instructional tasks by leveraging both textual plans and visual information. Specifically, we bring cross-modality through two key tasks: Conversational Video Moment Retrieval, where the model retrieves relevant step-video segments based on user queries, and Visually-Informed Step Generation, where the model generates the next step in a plan, conditioned on an image of the user's current progress. \modelname{} is trained using a novel multitask-multistage approach, designed to gradually expose the model to multimodal instructional-plans semantic layers, achieving strong performance on both multimodal and textual dialogue in a plan-grounded setting. Furthermore, we show that the model delivers cross-modal temporal and plan-structure representations aligned between textual plan steps and instructional video moments.~\footnote{The model, code, and non-personal data will be made publicly available at \url{https://github.com/dmgcsilva/mmplanllm}}
\end{abstract}

\section{Introduction}

\begin{figure}[ht]
    \centering
    \includegraphics[width=1.0\linewidth, trim={5mm 5mm 20mm 10mm}, clip]{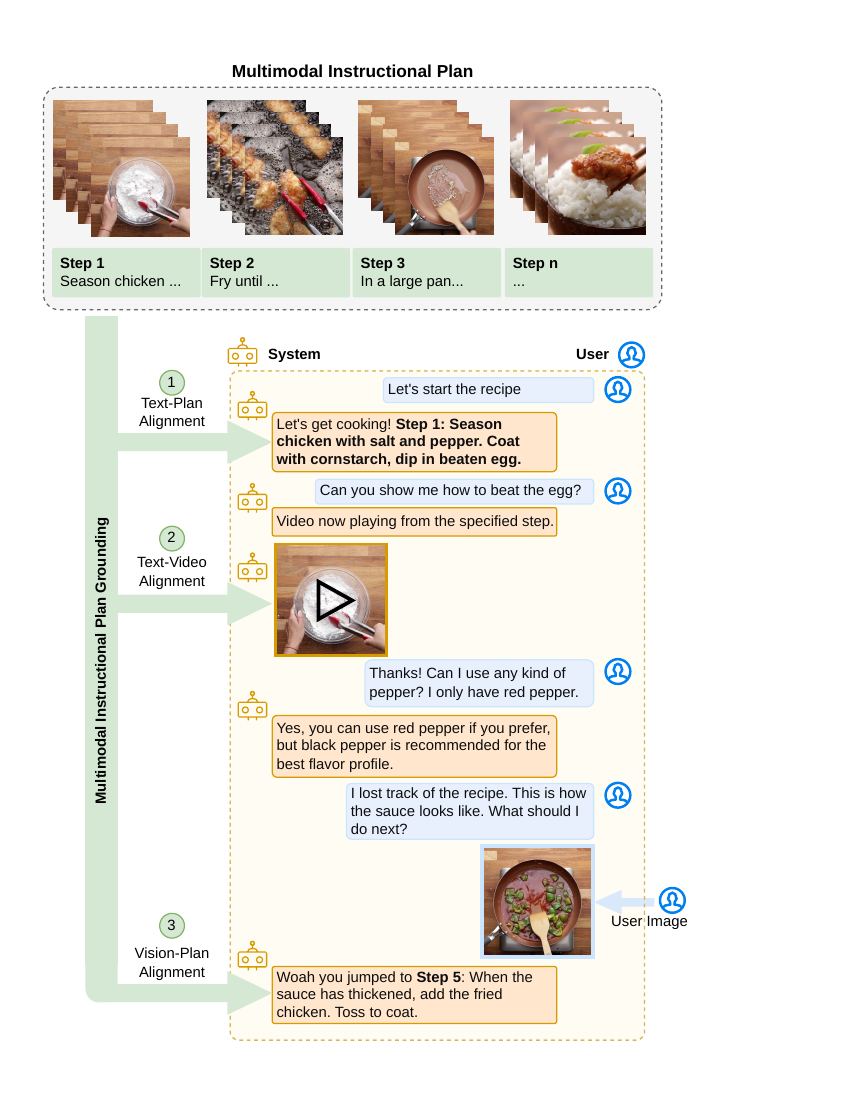}
    \caption{Example of a plan-grounded multimodal dialogue. The proposed model has the ability to understand and respond to multimodal input, provide relevant information from multiple knowledge sources, and guide the user through a complex task while adhering to a structured plan.}
    \label{fig:mmdial_subfig}
\end{figure}

The research of Large Language Models (LLMs) in assisting users with complex procedural plans, such as cooking or DIY projects, presents an exciting new frontier in NLP research~\cite{wizard_of_tasks}. 
However, while LLMs can excel at text-based conversational interactions~\cite{vicuna2023}, procedural plans are inherently multimodal, often accompanied by illustrative images or instructional videos~\cite{tasty_dataset, recipe1mplus}.
Thus, to interact in a reliable and trustworthy manner, it is crucial that these models not only comprehend procedural plans, but also ground the dialogue on these plans and align them with the visual domain, understanding images and videos to accurately assess progress and provide helpful guidance (Figure~\ref{fig:mmdial_subfig}).

In this work, we tackle this challenge and propose a multimodal LLM that is deeply grounded in both procedural text-plan and the accompanying visual-plan. Specifically, we focus on jointly learning three fundamental tasks: \textbf{Plan-following} capabilities where the LLM can generate and skip steps of the plan  (label 1 of Figure~\ref{fig:mmdial_subfig}), \textbf{Conversational Video Moment Retrieval} to retrieve a relevant step-video moment that accurately describes the current plan step (label 2 of Figure~\ref{fig:mmdial_subfig}), and \textbf{Visually-Informed Step Generation}, where, the goal is to, based on visual user input describing their current progress, generates the appropriate follow-up plan step (label 3 of Figure~\ref{fig:mmdial_subfig}).
To address these challenges, we propose \textbf{M}ulti\textbf{M}odal \textbf{Plan} \textbf{LLM} (\modelname{}), a dedicated model architecture capable of guiding users through a complex task plan, while supporting textual and visual plan information, both as input and output. In particular, we extend an LLM backbone with task-specific projection layers. These allow capturing video semantic and temporal information and supporting flexible decode-time multimodal retrieval, conditioned on task plans.
For training, we devise a novel multitask, multistage training approach designed to progressively instill the desired multimodal capabilities while preserving or improving on previously learned ones.

\modelname{}, the main contribution of this paper, is a model capable of guiding users through complex tasks, while adhering to the user requests, grounding the plan progress on user-uploaded images through visually-informed step retrieval, and performing conversational step-video moment retrieval. 
In particular, its groundbreaking multimodal plan-guiding capabilities,  lets it align image inputs with the correct step of the instructional plan, perform step-video moment retrieval, producing step-aligned cross-modal representations, with limited performance drop on text-only requests.

A thorough evaluation shows \modelname{}'s competitive performance on text-only tasks against task-specific baselines, and substantial improvements over existing approaches on multimodal tasks.

\section{Related Work}

In recent years, with the release of large open source foundational models such as OPT~\cite{opt}, Llama~\cite{llama} and others~\cite{gpt2, gpt3, mistral7b}, the field of Large Language Models (LLMs) for conversational settings has received significant attention. Due to this, the contributions have been diverse, with work focusing on improving training data~\cite{vicuna2023,llama2}, scaling model size~\cite{chowdhery2022palm}, and adopting a Mixture of Experts (MoE) architecture~\cite{mixtral, shen2023mixtureofexperts}.
The applications of these models are varied, such as instruction following~\cite{instructgpt, alpaca, instructions_mishra}, conversational dialogue~\cite{Zhang2019DIALOGPTL, vicuna2023}, and other task-specific applications~\cite{t5, simpletod}.

Researching models capable of understanding multimodal input has also been a topic of great interest. 
A common approach has been the usage of pretrained LLMs and Visual Encoders to achieve efficient and effective Large Vision-Language Models (LVLMs) with limited resources; however, the way these models interface has been varied. Some approaches, such as the LLaVa models~\cite{llava, llava_rlhf} and FROMAGe~\cite{fromage} have found that linear projections are enough. Others deploy larger "interpretation" modules such as the Q-Fromer in BLIP~\cite{blip_2, instructblip}, the Visual Abstractor in mPLUG-Owl~\cite{mplug_owl}, or the Perceiver~\cite{perceiver} employed in Flamingo~\cite{flamingo}. Another interesting approach is the modification of the internal Transformer~\cite{vaswani} attention blocks 
such as the visual expert in CogVLM~\cite{cogvlm} and the Modality-Adaptive Module in mPLUG-Owl2~\cite{mplug_owl2}.
Some work has also been done on training multimodal models from scratch such as PaLi~\cite{pali}, Gemini~\cite{geminiteam2023gemini}, and Large World Model~\cite{lwm}.

Video Moment Retrieval (VMR) is the task of, given a video and textual prompt that describes an action or event that occurs in a video, retrieving a video clip from within said video that best matches the provided textual prompt. Proposal-driven approaches focus on identifying candidate proposals and then ranking them to find the most relevant one~\cite{Gao2021-iw, Wang2022-uf, role_vmr}. In contrast, others opt for a proposal-free approach that predicts the target moment directly from the video-prompt feature mappings~\cite{siamese_vmr, zhang-etal-2020-span, Yuan2018ToFW} often relying on cross-modal attention modules or on learnable query embeddings such as EaTR~\cite{Jang_2023_ICCV} or MH-DETR~\cite{Xu2023MHDETRVM}. 
A common problem in VMR is the need to do extensive and expensive temporal annotations, 
an alternative is point-level VMR where the annotation is a single frame point~\cite{jiang_pointlevel} or a small segment~\cite{partial_vmr}.
Recently, several approaches have been adopting a Detection Transformers~\cite{detr}, as it does away with the need for many hand-designed components, and tackling the problem as a direct set prediction~\cite{Lei2021QVHighlightsDM, Sun2024TRDETRTT, Moon2023QueryD, Lei2021DetectingMA}.

\section{Multimodal Plan-Grounded LM}
In this section, we present the main elements of this work:
we start by formalizing the problem, then we describe \modelname{}, its architecture, and the multi-stage training process used. We end by detailing how the supporting synthetic training dataset is generated.

\subsection{Problem Definition}

Let $D = \left\langle P, T, V \right\rangle$ be a dialogue that consists of a procedural plan $P$ composed of $k$ sequential steps $P = \{ s_1, \cdots , s_k\}$, and a set of $n$ user-system interaction turns $T = \left\{ t_1, \cdots, t_n \right\}$, where a turn $t_i = \left\langle U_i, R_i, I_i^* \right\rangle$ is composed of a user request $U_i$, a system response $R_i$ and, optionally, a user-uploaded image $I_i$, and $V$ a video that demonstrates how to follow the plan $P$. $V$ is composed of $i$ frames $V = \{ f_1, \cdots, f_l \}$.
Here, a plan step is a sequence of words $s = \{ w_1, w_2, \cdots \}$, and a video moment $m_V$ is the sequence of video frames denoted by its starting and ending frame $m_V = \{ f_s, f_e \}$ with $f_s, f_e \in V$ and $s, e \le l$. Each video moment represents a plan step or part of it.

Based on the user-request type, our approach simultaneously adapts and performs interleaved multimodal plan-grounded tasks. In particular, three key features are supported: general plan-grounded answer generation, conversational video moment retrieval, and visually-informed step generation.
These key features are delivered by extending a vision and language model, in a multi-task setting, through a multi-stage training scheme.

\subsection{~\modelname{} Learning}
\paragraph{Plan-Grounded Answer Generation (PGAG).}
In this task, given a dialogue $D_j$ and the latest user request $U_{i+1}$ the objective is to generate $R_{i+1}=\{w^r_1,\ldots,w_n^r\}$ that adequately answers the user request, while conditioning on the previous turns $T_{i-c:i}$, with $c$ being the context size and $1\leq i<i+1$. The objective is formulated as a plan-grounded cross-entropy loss,
\begin{align*}
    \mathcal{L}_{pgag} = -\sum_{t=1}^T \log P(w_t^r|w^r_{1:t-1}, \bm{U_{i+1}}, D_j)
\end{align*}

\paragraph{Conversational Video Moment Retrieval (CVMR).}
This task seeks to retrieve a video moment that illustrates the current step of the task plan. Namely, given a textual user video request $U_{i+1}$, it seeks to retrieve the relevant video moment from a video $V$, given a dialogue $D_j$, considering only the previous turns $T_{i-c:i}$, with $c$ being the context size. To formulate the retrieval problem, \modelname{} generates a system response $R_{i+1}$ and locates the corresponding video moment $m_V$ within $V$. For tractability, we focus on retrieving a single keyframe $f_m$ that represents moment $m_v$. We define $f_m$ as the relevant segment's middle frame, with $m = \lfloor \frac{e - s}{2} \rceil$.
Recognizing the high similarity between consecutive frames (see Appendix~\ref{sec:app_framesim}), we formulate a video moment retrieval task by relaxing the retrieval target to consider a bidirectional context window of $N$ adjacent frames. This translates to retrieving any frame in a window of $2N+1$ frames centered around $f_m$. Specifically, the two-component loss is formulated as follows:
\begin{align*}
\begin{split}
    &\mathcal{L}_{ret}=\sum_{k=m-N}^{m+N} \frac{-\text{log}\left(P(f_k|D_j, U_{i+1})\right)}{2N+1}  \\
    &\mathcal{L}_{cvmr} = \mathcal{L}_{ret} + \mathcal{L}_{pgag}
\end{split}
\end{align*}

\paragraph{Visually-Informed Step Generation (VSG).}
In this last task, given a user request $U_{i+1}$ and a user-uploaded image $I_{i+1}$, that visually depicts their current progress on the task being executed, the goal is to generate an appropriate system response $R_{i+1}$ , that accurately copies the relevant plan step $s$, while accounting for the conversational history $D_j$, considering only the previous turns $T_{i-c:i}$, with $c$ being the context size. The loss is formulated as a visually conditioned cross-entropy loss,
\begin{align*}
    \mathcal{L}_{vsg} = - \sum_{t=1}^{T} log~P(w_t^r|w_{t-1:1}^r,\bm{I_{i+1}},D_j, U_{i+1}).
\end{align*}

\subsection{Model Architecture}

\begin{figure*}[t]
    \centering
    \includegraphics[width=0.95\textwidth, trim={0mm 4mm 0mm 4mm}, clip]{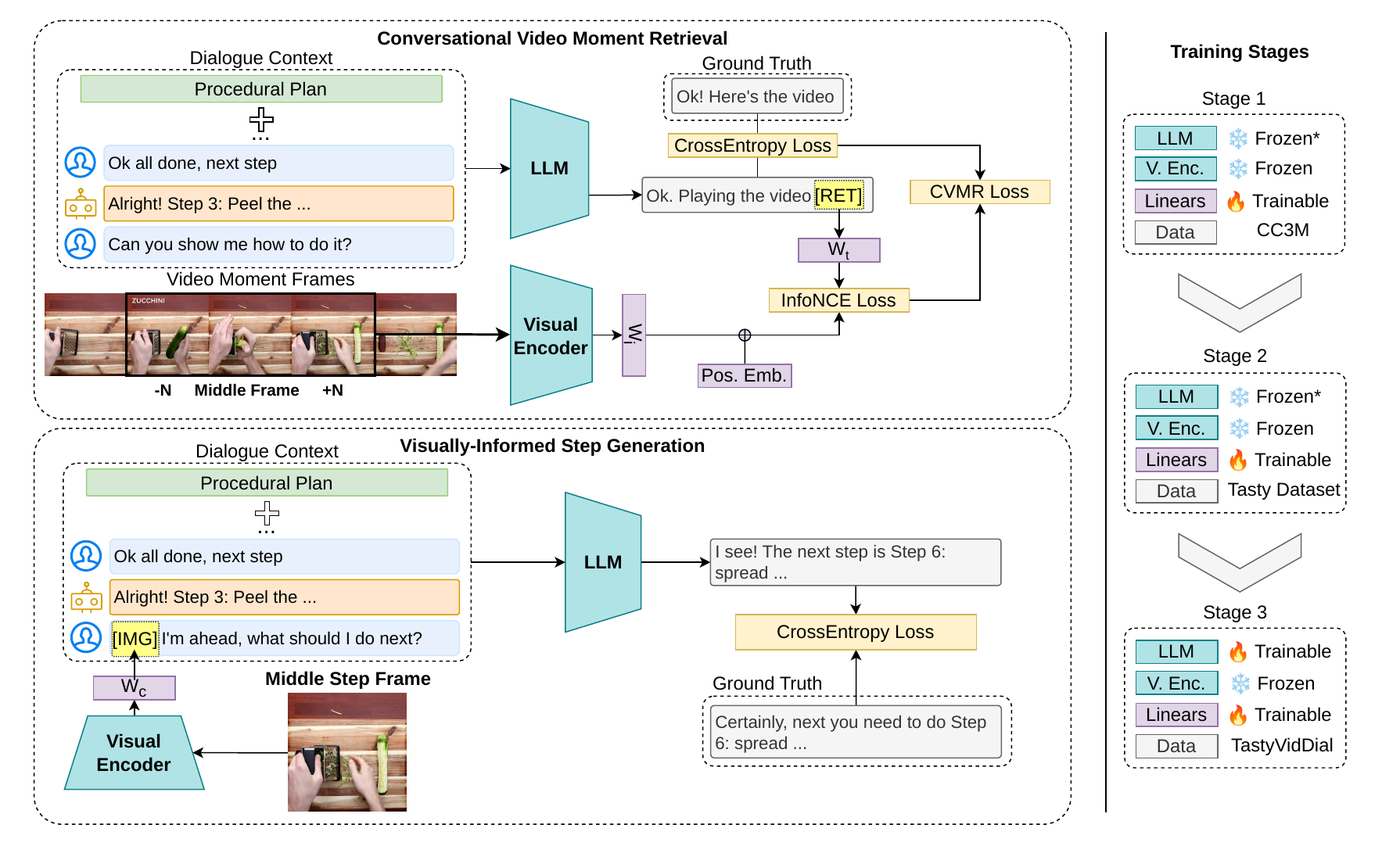}
    \caption{Comprehensive illustration of the \modelname{} architecture, including the 3 training stages employed for model training. *Denotes the \texttt{[RET]} token embedding representations and the Language Modeling Head of the LLM remain trainable.}
    \label{fig:planllmvis_arch}
\end{figure*}

The architecture of the proposed model, \modelname{}, expands on the framework presented in FROMAGe~\cite{fromage} and is composed of three main component groups: \textit{a) a language model backbone}, \textit{b) a vision encoder}, and \textit{c) task-specific projection layers}.
Each of these layers will be responsible for establishing an interface between the visual encoder and language model representations, while providing an efficient adaptation to new tasks, in a sequential or interleaved manner. 
Figure~\ref{fig:planllmvis_arch} provides an overview of this architecture. This section describes these three main component groups:

\paragraph{a) V\&L Model Backbone.}
The vision and language backbone model, takes as input a multimodal sequence, comprised of a user request $U_{n+1}$, the conversation history $D$, and an optional image $I_{n+1}$, and generates an appropriate system response.
For \modelname{}'s backbone model we use a pretrained decoder-only Transformer model. We experiment with different backbone models as detailed in Appendix~\ref{sec:app_ablation}.
The backbone LM model is trained with cross-entropy loss.

\paragraph{b) Video Encoder.}
Given a frame $I_n$ with resolution $H$x$W$, we leverage the ViT~\cite{vit} architecture, such that the video encoder outputs a learnable \texttt{[CLS]} token that attends to the entire frame, and a sequence of $N_v$ visual tokens $v_i \in \mathbb{R}^{d_{ve}}$, with $d_{ve}$ being the visual token embedding dimension. 
Each token is the result of attending to different non-overlapping patches of the frame, with $\frac{H}{N_v}\times\frac{W}{N_v}$ resolution. In \modelname{} this encoder remains frozen.

\paragraph{c) Task-specific layers.}
Support for novel tasks is achieved in \modelname{} through task-specific projection layers:
\begin{itemize}
    \item \noindent\textbf{VSG-specific layers.}
For VSG, and general Image-to-Text support,  we learn a single linear mapping  $W_c \in \mathbb{R}^{d_{ve} \times d}$, with $d$ being the LLM hidden dimension, that maps the \texttt{[CLS]}, obtained from the visual encoder, token to the embedding space of the language model, the resulting representation used to replace the \texttt{[IMG]} text embedding in the LLM.

\item \noindent\textbf{CVMR-specific layers.}
For CVMR, the model needs to be able to retrieve the middle frame of the relevant video clip, for the current moment.
In our task, each textual step is annotated with a relevant video segment.
The fact that these textual steps are not directly describing the clip visual content, but rather the actions that the user has to perform, poses a greater challenge, compared to traditional VMR datasets where the captions offer a visual description of the clip. 

We propose to address this challenge with a multi-stage multimodal plan-grounded training scheme, designed to close the visual$\Leftrightarrow$plan step semantic gap. Originally, a \texttt{[RET]} token is added to the language model's vocabulary, and is then appended to the end of each retrieval request. Its decoder-output embedding  is then mapped onto a cross-modal retrieval embedding space, using a trained linear mapping $W_t \in \mathbb{R}^{d\times q}$. A second linear layer is trained $W_i \in \mathbb{R}^{d_{ve}\times q}$ to map the visual features onto the retrieval space. We leverage this approach, and use the \texttt{[RET]} token to retrieve the video moment. 
For the training of these layers, we use the InfoNCE Loss~\cite{infonce} as the $\mathcal{L}_{ret}$ loss component, where we consider the middle clip frame, plus a bidirectional context-window of $N$ consecutive frames, as targets (i.e. positives). To incorporate temporal information we use fixed Rotary Positional Embeddings (RoPE)~\cite{rope} and apply temporal position shifting, where each positional embedding is shifted according to the frame's position within the video.
\end{itemize}

\subsection{Multi-stage Multimodal Training}
The model undergoes a multi-stage training scheme, in two core tasks: \textit{image captioning} and \textit{text-to-image retrieval}. We design a three-stage training approach tailored to our setting:

\paragraph{Stage 1. Visual Projection Layers.}
This preliminary phase is focused on bootstrapping the model's linear layers, $W_c$, $W_t$, and  $W_i$, by training on the \textit{image-captioning} and \textit{image-text retrieval} tasks.
For both tasks, we use the CC3M~\cite{cc3m} dataset, while the LLM and Visual Encoder are kept static. Only the embedding for the introduced \texttt{[RET]} token and the language modeling head are subject to training. 

\paragraph{Stage 2. Task Data Specialization.}
The subsequent stage seeks to specialize the model in the target domain. The same previous two proxy tasks are considered, but instead of general-domain data, we use domain-specific videos and captions. Specifically, we leverage the annotations present in the Tasty Dataset~\cite{tasty_dataset}. In this dataset, recipes are broken into actions, and these actions are then annotated with the start and end frame of the relevant video clip; we use these to create image-text pairs where the text is the action text and the image is the middle frame of the relevant clip. 

\paragraph{Stage 3. Multimodal Plan-Grounded Dialogue.}
The third, and most important, training stage aims to convey the necessary abilities to dialogue in the target plan-grounded dialogue setting on the recipes domain, attending to both uni- and multimodal user requests. To this extent,  plan-following multimodal instructional data is used (see section~\ref{sec:synth_dataset}), covering dialogue interactions, with particular emphasis on the envisioned multimodal interactions.
To facilitate training, we start with text-only samples and then move to multimodal ones, for the latter we alternate between CVMR and VSG batches. During this phase, the LLM is fully trained, along with all of the additional linear layers.

\subsection{Synthetic Multimodal Plan-oriented Training Data }
\label{sec:synth_dataset}

To prepare the model to cope with the wide range of user requests in plan-grounded dialogues, we resort to synthetic data generation. Namely, we build upon the methodology of PlanLLM, and further incorporate multimodal queries. This methodology follows a pipeline that utilizes real user-agent dialogues and, using an intent classifier, extracts a user policy and user utterances. To generate dialogues, user intents are selected for each turn, and a combination of templates, external knowledge bases, and generative models are used to create accurate system responses.
The incorporation of multimodal requests is accomplished by exploiting the Tasty Videos Dataset~\cite{tasty_dataset}, which comprises culinary recipes, each accompanied by a video and annotations delineating each step into individual actions and signaling the start and end of the said actions within the video. Herein, we detail how these annotations were leveraged to integrate multimodal user requests into the pre-existing data generation pipeline.

\paragraph{CVMR Requests.}
For the retrieval of specific video moments, the target clip corresponds to the one annotated for the current action. In instances where a plan step is composed of multiple actions, the first action is considered.

\paragraph{VSG Requests.}
Regarding VSG queries, a step subsequent to the user's current progress within the recipe is selected (e.g., if the user is at step 3, any step from 4 onwards is eligible), biasing to closer steps. The middle frame of the selected step is then used as the user-uploaded image that showcases the user's progress, at that point in the dialogue.

\vspace{2mm}
For both request types, the textual user requests and system responses are sampled from handwritten template lists. To improve diversity, an external generative model is prompted to extend the lists of possible user and system utterances. To this dataset we call \textbf{Tasty} \textbf{Vid}eo \textbf{Dial}ogue (\dataset{}).

\section{Experimental Setup}
\subsection{Instructional Tasks Datasets}
\label{sec:experimental_setup}
\paragraph{\dataset{}.} 
To conduct our experiments, we propose a novel dataset for conversational multimodal dialogue over complex tasks.
We create a dataset of 50k generated dialogues, between a user and a multimodal agent, while following complex tasks, resulting in $\approx 500k$ dialogue turns.
We utilize a set of 1500 illustrated recipes obtained from the Tasty Videos Dataset~\cite{tasty_dataset} to ground the generated dialogues.
To maximize dialogue quality, we only consider recipes with 5 to 10 steps, at least 6 ingredients, no more than 300 tokens, and at least 1 annotated video action for every step. To reduce frame count, we consider 1 for every 20 frames in the video.
For training, validation, and testing we use a 90/5/5 split.

\paragraph{Simulated Alexa TaskBot.} 
For the evaluation of text-only requests, we use the PlanLLM dataset as described in \citet{planllm}. We use this dataset version to avoid dialogue turns where the user request is text-only but one of the previous turns, present in the context, is multimodal.

\subsection{Methodology}

\paragraph{Backbone Models.} For the LM Backbone we use LLama2~\cite{llama2} (results with more models in Appendix~\ref{sec:app_ablation}). The visual encoder is  CLIP ViT-L/14~\cite{clip}. See Appendix~\ref{sec:app_details} for more implementation details.

\paragraph{Metrics.}
For evaluation of CVMR turns, we follow recent works~\cite{mmcda,Diwan2022-cp,Wang2022-uf} and use $R@n$, mean Average Precision (mAP),
Step Accuracy to measure if the retrieved frame is inside the video moment for the relevant step, and Mean Normalized Frame Distance (MNFD)
\begin{math}
    \mathrm{MNFD} = \frac{1}{N} \sum_{i=1}^{N} \frac{|f_{\text{retrieved},i} - f_{\text{target},i}|}{F_i},
    \label{eq:mnfd}
\end{math}
where $f_{retrieved,i}$ and $f_{target,i}$ are the retrieved and target frame respectively.

For the automatic evaluation of answer generation, we consider BERTScore(BS)~\cite{Zhang2019BERTScoreET} and ROUGE-L~\cite{Lin2004ROUGEAP}. 
To measure the VSG performance, apart from ROUGE-L, we use Exact Match, which measures whether the target step is contained, or not, in the system's response.

\paragraph{Protocol.}
Across all dialogue-based evaluations, we consider a context window of the 4 previous turns and pass the model the recipe steps along with the current step the user is on. The steps are included in the prompt in a numbered manner (eg, "Step 1 ..., Step 2 ...").
For CVMR, the model also sees the candidate system response and we extract the output embeddings for the position immediately after the generated \texttt{[RET]} token, if the model fails to generate the \texttt{[RET]} token we use the output embeddings of the first generated token. We set $N=2$.
For VSG and Answer Generation the model does not see any additional context. When evaluating a specific task, the model is not provided with any marker or information indicating the type of response wanted. For CVMR the candidate pool is composed of all of the frames of the recipe instructional video and the negative frames are the target frames for other samples in the same batch.

\paragraph{Baselines.}
As a baseline, we compare our approach with FROMAGe on a zero-shot setting, as it was not fine-tuned for our domain specifically. We also compare against a random baseline that randomly retrieves a frame from the video for CVMR, and, for VSG, it randomly selects a plan step from the ones not yet completed by the user.
For textual requests, we compare against the PlanLLM model, with no further fine-tuning, to gauge performance variance on text-only dialogue turns.

\section{Results and Discussion}
\label{sec:res_disc}

\subsection{Plan Grounding}
\begin{table}[t]
\centering
\small
\begin{tabular}{l|cc|cc} \toprule
\multirow{2}{*}{\textbf{Model}} & 
\multicolumn{2}{c}{\textbf{Answer Gen.}} & 
\multicolumn{2}{c}{\textbf{Plan-Navigation}} \\

& ROUGE & BS & Explicit & Implicit \\ \midrule
FROMAGe             & 29.98 & 63.55 & ---  & --- \\
PlanLLM             & \textbf{75.58}    & \textbf{88.66}    & \textbf{0.895}    & \textbf{0.480}    \\ \midrule
\modelname{}  & 66.58             & 83.28             & 0.855             & 0.440             \\ \bottomrule
\end{tabular}%
\caption{Instructional plan following generation results, on automatic metrics. PlanLLM results as reported in ~\cite{planllm} }
\label{tab:nav_eval}
\end{table}

\begin{table*}
\centering
\small
\begin{tabular}{l|l|cccccc|cc}
\toprule
\multirow{2}{*}{\textbf{Method}} &
\multirow{2}{*}{\begin{tabular}[c]{@{}c@{}}\textbf{LLM Backbone}\\ (\# Params)\end{tabular}} &
\multicolumn{6}{c|}{\textbf{Conversational Video Moment Retrieval}} &
\multicolumn{2}{c}{\textbf{VSG}} \\

& & \multicolumn{1}{c}{R@1} & 
\multicolumn{1}{c}{R@5} & 
\multicolumn{1}{c}{R@10} & 
\multicolumn{1}{c}{mAP} &
\multicolumn{1}{c}{Step Acc.} &
\multicolumn{1}{c|}{MNFD$\downarrow$} &
\multicolumn{1}{c}{Ex. Match} &
\multicolumn{1}{c}{ROUGE} \\ \midrule
Random & ---                & 1.65      & 8.66 & 17.10 & 7.70 & 16.12 & 32.21 & 28.02 & 37.51 \\ 
FROMAGe            & OPT (7B)  & 3.08 & 11.75 & 22.76 & 10.17 & 25.09 & 26.11 & 0.34 & 7.31 \\ 
MM-PlanLLM & Llama2 (7B)    & \textbf{5.50} & \textbf{38.53} & \textbf{53.82} & \textbf{21.52} & \textbf{54.10} & \textbf{13.26} & \textbf{38.16} & \textbf{42.62} \\ \bottomrule
\end{tabular}%
\caption{Evaluation results of our best-performing model \modelname{-Llama2}, on multimodal tasks, against the baselines. For the CVMR and VSG tasks we used the \dataset{} dataset.}
\label{tab:main_eval}
\end{table*}

\paragraph{Plan-Grounded Answer Generation.}
A key property of \modelname{} is its strong plan-following capabilities. 
To assess this, we evaluated \modelname{} on text-only dialogues, mirroring the main evaluation setting of PlanLLM. 
We utilized the original PlanLLM dataset, which exclusively comprises text-based conversations.

As shown in Table~\ref{tab:nav_eval}, \modelname{} achieves a BERTScore of 83.28, approximately 94\% of PlanLLM's performance (88.66), on answer generation in a text-only plan-grounded setting.
In contrast, FROMAGe demonstrates notably weaker performance in this setting.
To understand if this performance differential also reflects on \modelname{}'s ability to guide users through tasks, we replicated the GPT-4-based Plan Navigation evaluation from \citet{planllm}. The results in Table~\ref{tab:nav_eval} indicate that \modelname{} remains competitive on this task having 85.5 accuracy on explicit navigational requests, a 4.0 accuracy loss over PlanLLM, reinforcing that it retained the ability to effectively follow instructional plans.

\paragraph{Conversational Video Moment Retrieval.}

To assess CMVR performance, we evaluated \modelname{} on all video moment retrieval requests within the \dataset{} test set. We benchmarked \modelname{} against FROMAGe, which is capable of general conversational image retrieval, and the random baseline (described in Section~\ref{sec:experimental_setup}), to quantify the gains achieved through our task-specific training approach.

The results shown in Table~\ref{tab:main_eval} highlight the efficacy of our focused training.  \modelname{} significantly outperforms FROMAGe across all metrics, demonstrating over 100\% improvement in most cases. Whereas FROMAGe, in turn, demonstrates minimal improvement over the random baseline.
The performance gap between R@1 and R@5, coupled with a high Step Accuracy, suggests that while \modelname{} consistently identifies the relevant video moment (evidenced by high Step Accuracy), the high visual similarity between adjacent frames within the same video moment proves a challenge for R@1 scores. This is explored in Section~\ref{sec:alignment}.

\paragraph{Visually-Informed Step Generation.}
To evaluate \modelname{}'s ability to interpret visual input and align it with instructional plans, we also evaluated solely in the VSG requests.
Results for this task are shown in the second column group of Table~\ref{tab:main_eval}. There is a stark contrast in performance between \modelname{} and FROMAGe, with the latter rarely preserving the step text verbatim. Conversely, \modelname{} achieves an Exact Match score of 38\%.

\begin{figure}[t]
    \centering
    \includegraphics[width=\linewidth, trim={14mm 1mm 5mm 12mm}, clip]{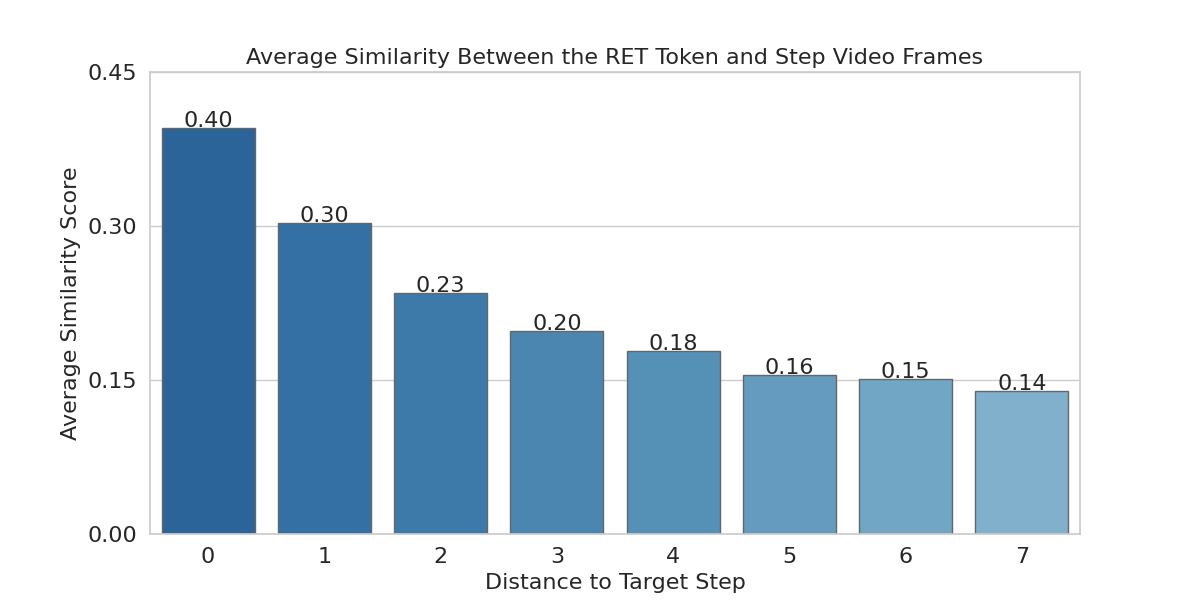}
    \caption{Text-query to visual plan alignment. \modelname{} effectively learns to align textual \texttt{[RET]} token representations with that of the target step frames. We remove outliers for clarity.}
    \label{fig:cvmr_sims}
\end{figure}

\begin{figure}[t]
    \centering
    \includegraphics[width=\linewidth, trim={14mm 1mm 5mm 12mm}, clip]{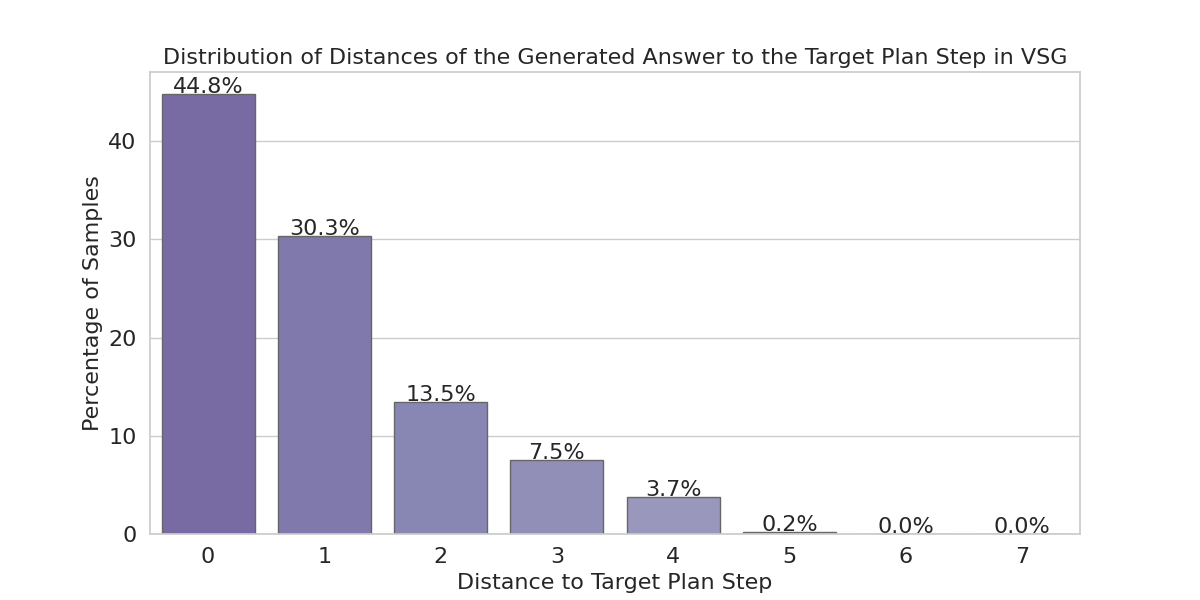}
    \caption{Image-query to text plan alignment. Most similar plan step to the provided visual input, as measured by BS using the generated answer.}
    \label{fig:vsg_dist}
\end{figure}

\begin{table*}[t]
\centering
\small
\begin{tabular}{l|cccccc|cc|cc}
\toprule
                    & \multicolumn{6}{c|}{\textbf{Conversational Video Moment Retrieval}}      & \multicolumn{2}{c|}{\textbf{VSG}} & \multicolumn{2}{c}{\textbf{Answer Gen.}} \\
                    & R@1  & R@5   & R@10  & mAP   & Step Acc. & MNFD$\downarrow$  & Ex. Match          & ROUGE         & ROUGE          & BS         \\ \midrule
LLaMa2 - Phase 1    & 2.05 & 12.13 & 19.02 & 9.01  & 16.42     & 29.25 & 0.00  & 6.48 & 31.78 & 63.44 \\
+ Phase 2           & 3.45 & 15.21 & 23.04 & 10.68 & 16.98     & 29.46 & 7.33  & 18.87 & 23.68 & 55.30 \\
+ Phase 3      & 6.72 & 35.26 & 48.69 & 20.80 & 52.52     & 14.03 & 37.14 & \textbf{42.84} & 66.11 & 83.03 \\ 
\quad + Adj. Frames & \textbf{7.46} & 30.50 & 48.60 & 20.43 & 52.05     & 14.14 & 34.58 & 40.59 & 65.78 & 82.95 \\
\quad + Pos. Embs.  & 5.50 & \textbf{38.53} & \textbf{53.82} & \textbf{21.52} & \textbf{54.10}     & \textbf{13.26} & \textbf{38.16} & 42.62 & \textbf{66.58} & \textbf{83.28} \\ \bottomrule
\end{tabular}
\caption{Impact of the several training stages on model performance on the three main tasks.}
\label{tab:abalation_phases}
\end{table*}

\subsection{Multimodal Plan Alignment}
\label{sec:alignment}
\paragraph{Text to Visual Plan.}
In Figure~\ref{fig:cvmr_sims} we plot the average similarity between the \texttt{[RET]} token and all frames of each plan step in the plan video, ordered by their absolute distance to the target step.
The results demonstrate that \modelname{} effectively learns to produce a representation that aligns closely with the video moment relevant to the target step. This is supported by the significantly higher similarity scores observed for frames within the target step (distance 0) compared to frames from other steps. Additionally, the gradual decline in similarity as the distance from the target step increases, further confirms the model's ability to discriminate between relevant and irrelevant video moment frames based on the textual plan.

\paragraph{Image to Text Plan.}

To assess \modelname{}'s ability to align visual representations of steps with the corresponding textual descriptions, we used BERTScore to measure the similarity between generated answers and plan steps in the VSG task. Then, for each VSG instance, we identified the plan step with the highest BERTScore similarity to the generated answer and plotted its distance from the actual target step in Figure~\ref{fig:vsg_dist}.

The distribution in Figure~\ref{fig:vsg_dist} reinforces that \modelname{} demonstrates a substantial capacity for aligning visual input with the corresponding textual step, achieving a success rate of 44.8\% on the test set.
Moreover, 30.3\% generated answers are most similar to steps immediately preceding or following the target step, highlighting the model's ability to capture the sequential nature of instructions and identify steps closely related to the visual input.

\subsection{Ablation Study}
\label{sec:ablation}
We conducted ablation studies to investigate the impact of each training stage and architectural choices on model performance across all tasks. 

\paragraph{Training Stages.}
To train \modelname{}, we devised a multi-stage approach to maximize performance gains and minimize catastrophic forgetting. This analysis can be seen in Table~\ref{tab:abalation_phases}.
These results show that each of the three training stages contributed incrementally to improving the targeted capabilities. Stage 1, which focused on general image understanding, established a foundation for the model to outperform the random baseline in CVMR. Stage 2, which aimed to instill domain-specific multimodal understanding, further enhanced performance on both CVMR and Step Generation tasks, even before explicitly training on these tasks. Finally, Stage 3, where we integrated conversational abilities, led to substantial improvements across all three tasks, highlighting the importance of end-to-end task-specific training.

Within the last stage, we also report the improvements provided both by the usage of adjacent frames as candidates and usage of positional embeddings for CVMR training.
Surprisingly, utilizing multiple candidate frames for CVMR training yielded minimal benefits. We hypothesize that this is due to the high similarity between consecutive frames in video moments. However, the addition of positional embeddings, which incorporate temporal information, significantly improved performance across the board, underscoring the model's ability to leverage this additional context.

\paragraph{LLM Backbone.}
To understand how different LLM Backbones affect model performance we evaluated 8 LLMs on CVMR, VSG, and PGAG. The evaluated models ranged in size from 1.8B (Qwen-1.5~\cite{bai2023qwen}) to 7B parameters (Vicuna1.5~\cite{vicuna2023}). We consider PlanLLM as a backbone but skip the text-only data training in Stage 3.
On the CVMR evaluation Llama2 achieves the best performance across most metrics, with the exception of R@1 that is led by PlanLLM with 6.16 R@1. For VSG, PlanLLM exhibited a substantial lead over Vicuna in Exact Match (42.76 vs. 40.37), while Mistral \cite{mistral7b} achieved the highest ROUGE score (46.60).

In the answer generation task, most models demonstrated similar performance.  Our approach generalizes well for smaller LLMs, such as Qwen-1.5 (1.8B Param.) and Phi-2 (2.7B Param.) which achieved, on average, 88\% and 95\% of Llama2's performance, respectively.
The complete results are shown in the Appendix~\ref{sec:app_ablation}.

\section{Conclusion}
We propose \modelname{}, a multimodal architecture that enables multimodal comprehension for LMs in plan-grounded conversational settings. 
We follow a multistage training paradigm, coupled with task-specific synthetic data creation, that enables the model to slowly acquire the necessary abilities to understand multimodal input and generate multimodal outputs.

Experimental results demonstrates that \modelname{} outperforms task-specific baselines, showcasing minimal performance loss in text-only dialogues, while being capable of aligning textual steps with video moments and user images with the plan steps. The ablation study further highlights the effectiveness of the multi-phase training methodology and the value of incorporating temporal information. 

\section*{Limitations}
While \modelname{} addresses two key multimodal request types (CVMR and VSG) crucial for plan-grounded dialogue, we acknowledge that a complete system would need broader multimodal support, including visual question answering. 
Furthermore, long-term dialogue dependencies remain a challenge due to the limited context window of 4 turns during training (limited by the available hardware), hindering the model's ability to effectively recall and utilize information from earlier turns in the conversation. This limitation may impact the model's performance in extended interactions where maintaining context is essential.
We plan to address these limitations in future work.

\section*{Acknowledgements}
This work was supported by the FCT Ph.D. scholarship grant Ref. PRT/BD/152810/2021 awarded by CMU Portugal Affiliated Ph.D. program, and by the FCT project NOVA LINCS Ref. (UIDB/04516/2020). Data collection was possible under the Alexa Prize Taskbot Challenge organized by Amazon Science.

\bibliography{anthology,custom}
\bibliographystyle{acl_natbib}

\appendix

\section{Implementation Details}
\label{sec:app_details}

Table~\ref{tab:app_hyperparameters} details some of the hyperparameters used.
Each model is trained for 10k, 5k, and 2k steps for each phase, using a batch size of 64 (and 16 on multimodal batches in phase 3) on a single A100 40GB GPU. For optimization, we use the AdamW optimizer~\cite{adamw} with $\beta1 = 0.9$, $\beta2 = 0.95$, and $\epsilon = 1*10^{-5}$ for all runs. We used a constant learning rate of $1*10^{-5}$, for the third stage, with no warmup steps. All images are resized to fit a 224x224 image resolution. For phase 3, text-only training was separated from the multimodal training with the first 1k steps being text-only and the later 1k being multimodal.
The visual encoder used was CLIP ViT-L/14~\cite{clip}, the retrieval embedding dimension was set to 512, and the embedding dimension was kept the same as the LM Backbone so it varied from model to model.
For the 3rd phase we use LoRa~\cite{hu2022lora}, when training the LM Backbone, with a $r = 4$ and $\alpha = 8$ to reduce memory requirements. 

For BERTScore calculations we utilize \texttt{microsoft/deberta-xlarge-mnli}.

\begin{table}[]
\centering
\resizebox{\linewidth}{!}{%
\begin{tabular}{l|ccc}
\toprule
\textbf{Stage}  & \textbf{1}  & \textbf{2}  & \textbf{3}  \\ \midrule
Batch Size      & 64          & 48          & 1/4           \\
Grad. Acc.      & 64          & 1           & 64/4           \\
Train Steps     & 10000       & 5000        & 2000        \\
Val. Freq.      & 1000        & 1000        & 500         \\
GPU \#          & 1           & 1           & 1           \\
Seq. Max Len.   & 24          & 45          & 800         \\
DType           & BF16        & BF16        & BF16        \\ \midrule
Learning Rate   & $5*10^{-4}$ & $1*10^{-4}$ & $5*10^{-4}$ \\
Scheduler       & Constant    & Constant    & Constant    \\
Optimizer       & AdamW       & AdamW       & AdamW       \\
T. Emb. Dropout & 0.1         & 0.1         & 0.1         \\
Ret. Dimension  & 512         & 512         & 512         \\ \midrule
LoRa DType      & ---         & ---         & 16 bits     \\
LoRa Rank       & ---         & ---         & 4           \\
LoRa $\alpha$   & ---         & ---         & 8           \\
LoRa Dropout    & ---         & ---         & 0.1         \\ \bottomrule
\end{tabular}}
\caption{Hyperparameters used to train \modelname{} models across all three stages.}
\label{tab:app_hyperparameters}
\end{table}

\section{LM Backbone Ablation}
\label{sec:app_ablation}

We consider a comprehensive array of language model backbones in order to assess their impact on the overall model and select the best-performing one. In particular, we consider \textbf{Qwen-1.5}~\cite{bai2023qwen}, \textbf{Phi-2}, \textbf{Gemma2b}~\cite{gemma}, \textbf{Mistral-v0.1}~\cite{mistral7b}, \textbf{OPT}~\cite{opt}, \textbf{PlanLLM}~\cite{planllm}, \textbf{LLama2}~\cite{llama2}, and \textbf{Vicuna-7B}~\cite{vicuna2023}. As such, we cover LM backbones of different sizes, pre-training, and fine-tuning schemes.

\begin{table*}
\centering
\small
\begin{tabular}{l|cccccc|cc|cc}
\toprule
\multirow{2}{*}{\begin{tabular}[c]{@{}c@{}}\textbf{LM Backbone}\\ (\# Params)\end{tabular}} &
\multicolumn{6}{c|}{\textbf{Conversational Video Moment Retrieval}} &
\multicolumn{2}{c|}{\textbf{VSG}} &
\multicolumn{2}{c}{\textbf{Answer Gen.}} \\

& \multicolumn{1}{c}{R@1} & 
\multicolumn{1}{c}{R@5} & 
\multicolumn{1}{c}{R@10} & 
\multicolumn{1}{c}{mAP} &
\multicolumn{1}{c}{Step Acc.} &
\multicolumn{1}{c|}{MNFD$\downarrow$} &
\multicolumn{1}{c}{Ex. Match} &
\multicolumn{1}{c|}{ROUGE} &
\multicolumn{1}{c}{ROUGE} &
\multicolumn{1}{c}{BS} \\ \midrule
Qwen-1.5 (1.8B)   & 4.2  & 27.52 & 46.83 & 17.11 & 44.68 & 15.66 & 39.18 & 44.19 & 64.21 & 81.63 \\
Gemma (2.5B)      & 3.08 & 24.44 & 47.39 & 15.69 & 46.46 & 15.45 & 37.14 & 42.30 & 63.33 & 82.07 \\
Phi-2 (2.7B)      & 6.06 & 31.53 & 52.05 & 18.69 & 53.26 & 13.73 & 39.18 & 43.44 & 54.35 & 77.97 \\
OPT (7B)          & 4.48 & 31.44 & 52.61 & 18.80 & 50.84 & 15.34 & 35.78 & 43.15 & 38.52 & 70.65 \\
Mistral-v0.1 (7B) & 5.69 & 33.40 & 50.56 & 19.70 & 47.95 & 14.51 & 39.52 & \textbf{46.60} & 61.47 & 80.81 \\
PlanLLM (7B)      & \textbf{6.16} & 33.30 & 52.33 & 20.20 & 44.68 & 14.06 & \textbf{42.76} & 44.67 & 66.64 & 83.13 \\
Llama2 (7B)       & 5.50 & \textbf{38.53} & \textbf{53.82} & \textbf{21.52} & \textbf{54.10} & \textbf{13.26} & 38.16 & 42.62 & 66.58 & 83.28 \\
Vicuna1.5 (7B)    & 6.06 & 32.93 & 50.84 & 20.02 & 53.64 & 13.66 & 40.37 & 43.38 & \textbf{68.13} & \textbf{84.05} \\ \bottomrule
\end{tabular}%
\caption{Evaluation results of different LM Backbones for multimodal requests in the \dataset{} dataset.}
\label{tab:ablation_models}
\end{table*}

While we report our main evaluation results in Section~\ref{sec:res_disc} using Llama2~\cite{llama2} as the LM backbone, we trained a total of 8 models by varying the LM backbone. This sought to not only assert which was the best-performing model but also understand the impact of scaling the LM backbone on all tasks.

The results from this analysis, shown in Table~\ref{tab:ablation_models}, show a surprisingly low performance differential between models for all three tasks, with the only clear outlier being OPT~\cite{opt} on the answer generation task.
For Conversational Video Moment Retrieval we see a clear lead for Llama2~\cite{llama2} for most metrics, particularly for R@5, and a close second for Step Accuracy.
For Step Generation, PlanLLM~\cite{planllm} outperforms the other models on Exact Match whereas Mistral holds a small lead on BertScore. On this task Llama2 underperforms indicating that there might be a performance tradeoff between this task and the previous. 
For Answer Generation Vicuna performs the best likely due to its pretrain in a conversational setting, despite this both Llama2 and PlanLLM also perform closely to Vicuna. 
Focusing on \modelname{PlanLLM}, on phase 3 we skipped training on text-only samples as the model already had been trained on this setting, despite this the model is still competitive across all 3 tasks showing that our training approach seems to be agnostic to the models' pertaining tasks.

\section{Frame Similarity}
\label{sec:app_framesim}

\begin{figure}
    \centering
    \includegraphics[width=\linewidth, trim={20mm 9mm 32mm 15mm}, clip]{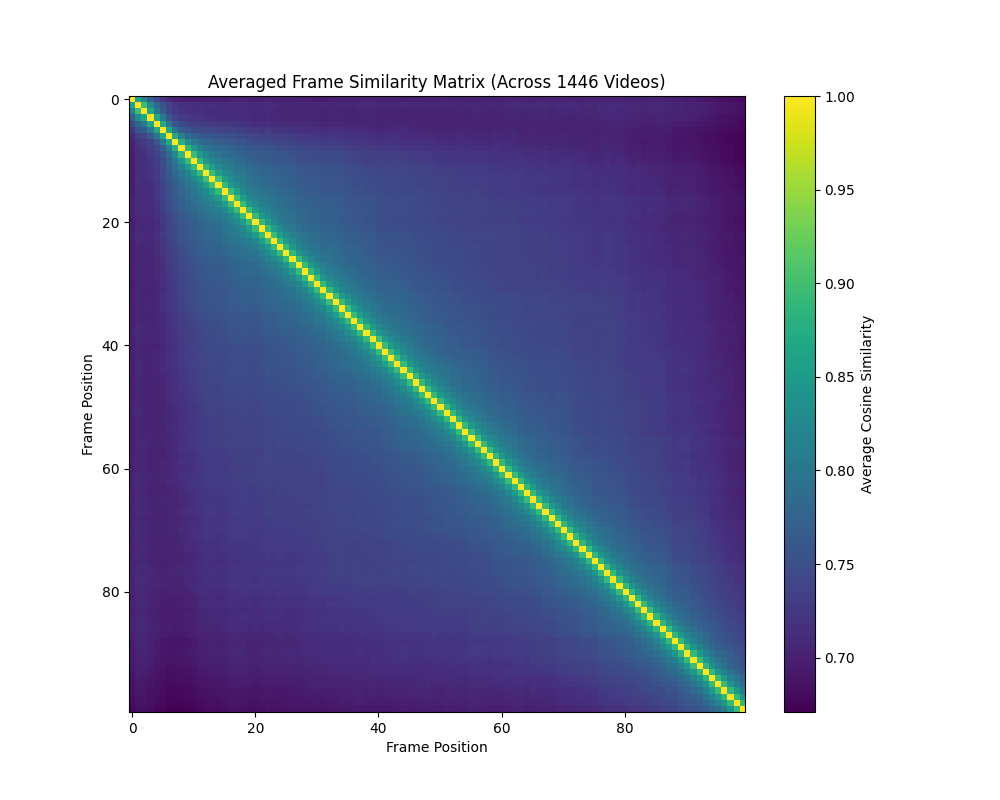}
    \caption{Average similarity of each frame against all other frames from the same video. It shows a clear bidirectional 3-frame window of higher similarity. }
    \label{fig:framesim_map}
\end{figure}

To investigate the degree of visual similarity between frames within recipe videos, we conducted an analysis using a subset of 1446 recipe videos from our dataset, each containing 100 or more frames. For each frame within the first 100 frames of a video, we computed its cosine similarity with all other frames in the same video using a CLIP image encoder, and averaged the similarity for each frame position across all videos.

The resulting averaged similarity matrix, shown in Figure~\ref{fig:framesim_map}, confirms that frames exhibit exceptionally high similarity to their immediate neighbors (mostly on a bidirectional 3 to 5-frame window), with a gradual drop-off in similarity beyond that point. Interestingly, we also note how similar frames from the same video tend to be with most frames having at least 0.7 similarity score to every other frame in the video. This underscores the need of visual encoders capable of differentiating the subtle visual changes that separate frames relevant to different plan steps.

\section{CVMR and VSG Examples}
\label{sec:app_examples}

\begin{figure*}[t]
    \centering
    \includegraphics[width=\textwidth]{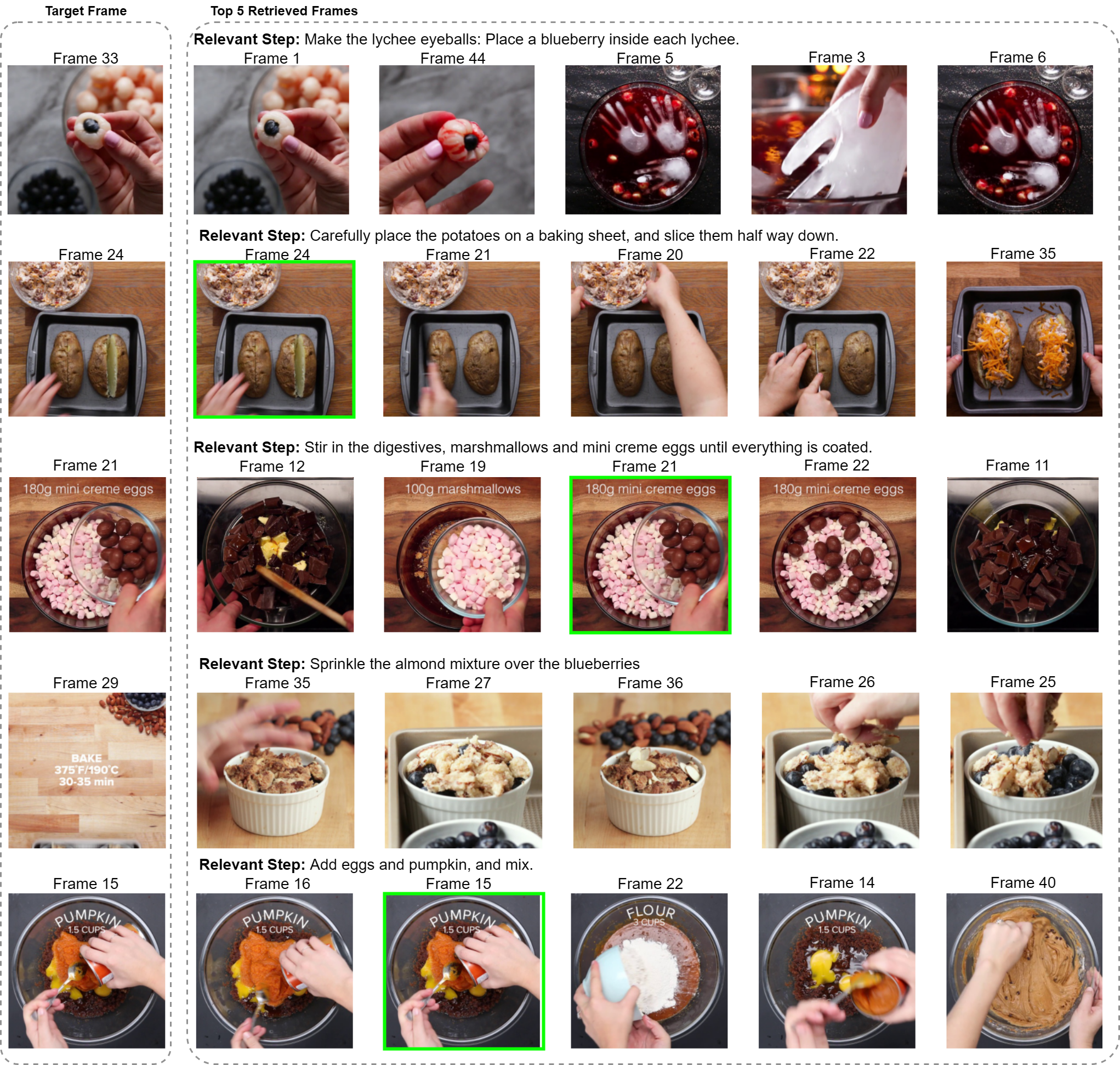}
    \caption{Five examples of CVMR results from the \dataset{} test set. These examples demonstrate that the model is adept at identifying the key elements that should be characterized in the target frame. However, in some cases, the selected middle frame may not be ideal (3rd and 4th rows), and in others, alternative frames may more accurately represent the target plan step (1st row). The conversational dialogue context has been omitted for readability, and green boxes highlight the target frame amongst the retrieved ones.}
    \label{fig:cvmr_5_examples}
\end{figure*}

\begin{figure*}
    \centering
    \includegraphics[width=\textwidth, trim=0mm 60mm 0mm 0mm,clip]{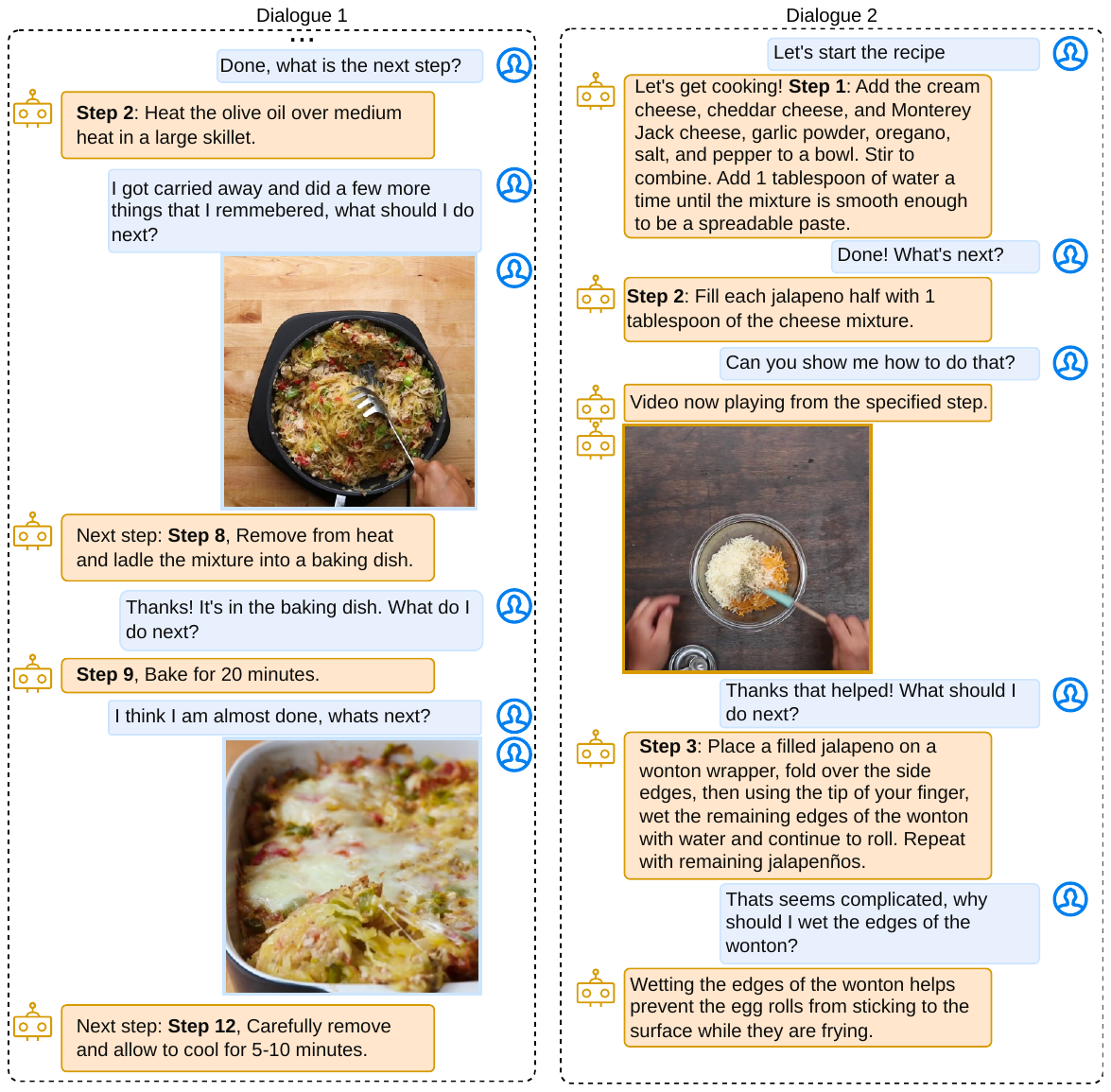}
    \caption{Real multimodal dialogues carried out by a volunteer interacting with \modelname{}. These dialogues showcase the model's ability to carry out full conversations with interleaved multimodal requests, without hindering text-only performance.}
    \label{fig:mm_dialogues}
\end{figure*}

In this section, we include a few examples of CVMR (Figure~\ref{fig:cvmr_5_examples}) and VSG (Table~\ref{tab:app_vsg_examples}) generations, extracted from the dataset test set. Additionally, in Figure~\ref{fig:mm_dialogues}, we showcase two dialogues collected by having a volunteer interact with the system. These examples and dialogues, demonstrate the model's performance on both single and multi-turn scenarios, showing that it can accurately answer to a wide range of requests in the target setting.

\begin{table*}[t]
\centering
\begin{tabular}{l|l|p{0.65\linewidth}}
\toprule
\textbf{User Image} & & \\ \midrule
\multirow{2}{*}{\raisebox{-\height}{\includegraphics[width=0.2\textwidth]{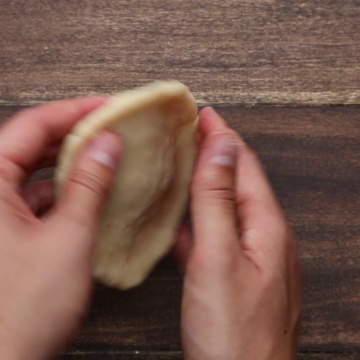}}} & \textbf{Target:} & Step 5: Pinch a piece of dough, about the size of a large egg, and roll into a ball. Using your hands, flatten the dough and spoon filling into the center. Wrap the dough into a pear shape and make sure there aren’t any holes. \\[1ex]
& \textbf{Gen:} & Continue with Step 5, Pinch a piece of dough, about the size of a large egg, and roll into a ball. Using your hands, flatten the dough and spoon filling into the center. Wrap the dough into a pear shape and make sure there aren’t any \\ \midrule

\multirow{2}{*}{\raisebox{-0.85\height}{\includegraphics[width=0.2\textwidth]{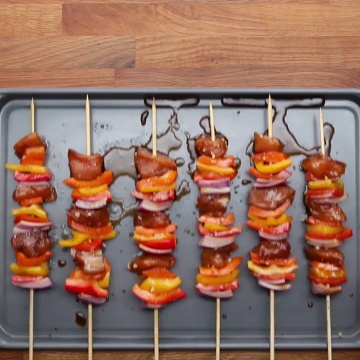}}} & \textbf{Target:} & Step 8: Bake or grill for 20-30 minutes or until chicken is no longer pink. \\[4ex]
& \textbf{Gen:} & Continue with Step 8, Bake or grill for 20-30 minutes or until chicken is no longer pink. \\[4ex]  \midrule

\multirow{2}{*}{\raisebox{-0.85\height}{\includegraphics[width=0.2\textwidth]{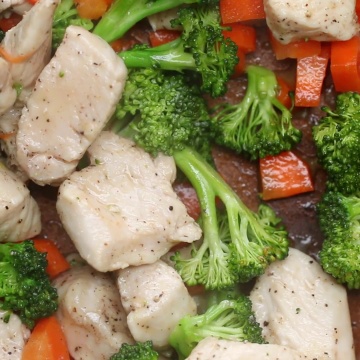}}} & \textbf{Target:} & Step 4: Add the rice, soy sauce, and peas, and mix thoroughly. \\[5ex]
& \textbf{Gen:} &  Continue with Step 4, Add the rice, soy sauce, and peas, and mix thoroughly. \\[5ex]  \midrule

\multirow{2}{*}{\raisebox{-\height}{\includegraphics[width=0.2\textwidth]{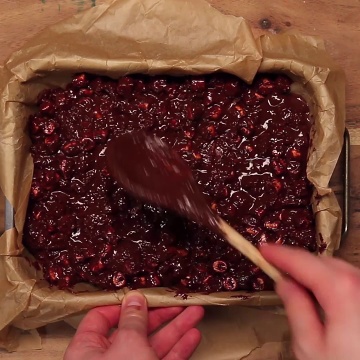}}} & \textbf{Target:} & Step 4: Melt 75g of the white chocolate and stir in some of the yellow food colouring until the desired colour is reached. If it seizes on you, add a little bit of oil and stir it until it reaches a runny consistency again. \\
& \textbf{Gen:} & Continue with Step 4, Melt 75g of the white chocolate and stir in some of the yellow food colouring until the desired colour is reached. If it seizes on you, add a little bit of oil and stir it until it reaches a runny consistency again. \\ \midrule

\multirow{2}{*}{\raisebox{-0.80\height}{\includegraphics[width=0.2\textwidth]{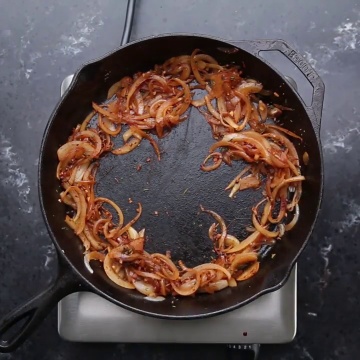}}} & \textbf{Target:} & Step 5: Add the shrimp and let them cook for 3 minutes. Flip and cook for another 3 minutes. \\[4ex] 
& \textbf{Gen:} & Continue with Step 4, Add garlic, rosemary, thyme, smoked paprika, salt and pepper. Stir and cook an additional 2 minutes. \\[4ex] 
\bottomrule
\end{tabular}
\caption{Five examples showcasing Visually-Informed Step Generation, from the test set. The model successfully identifies the appropriate next step in most cases, leveraging dialogue history and user-uploaded images. However, the difficulty increases when uploaded images lack clarity regarding the user's progress, as seen in the final example where it is unclear if the ingredient from Step 4 have already been added. Dialogue history ommited for readability.}
\label{tab:app_vsg_examples}
\end{table*}


\end{document}